\documentclass[11pt,letterpaper]{article}
\usepackage{emnlp2017}
\usepackage{times}
\usepackage{latexsym}
\usepackage{graphicx}
\usepackage{float} 

\emnlpfinalcopy

\title{Character-level Intra Attention Network for Natural Language Inference }

\author{Han Yang \and Marta R. Costa{-}juss{\`{a}} \and Jos{\'{e}} A. R. Fonollosa \\
       TALP Research Center \\
       Universitat Polit\`{e}cnica de Catalunya\\ 
  {\tt han.yang@est.fib.upc.edu  \{marta.ruiz,jose.fonollosa\}@upc.edu}}

\date{}

\begin{document}

\maketitle

\begin{abstract}
Natural language inference (NLI) is a central problem in language understanding. End-to-end artificial neural networks have reached state-of-the-art performance in NLI field recently. 

In this paper, we propose Character-level Intra Attention Network (CIAN) for the NLI task. In our model, we use the character-level convolutional network to replace the standard word embedding layer, and we use the intra attention to capture the intra-sentence semantics. The proposed CIAN model provides improved results based on a newly published MNLI corpus.
\end{abstract}

\section{Introduction}
Natural language inference in natural language processing refers to the problem of determining a directional relation between two text fragments. Given a sentence pair (premise, hypothesis), the task is to predict whether hypothesis is entailed by premise, hypothesis is contradicted to premise, or whether the relation between premise and hypothesis is neutral.

Recently, the dominating trend of works in natural language processing is based on artificial neural networks, which aims at building deep and complex encoder to transform a sentence into encoded vectors. For instance, there are recurrent neural network (RNN) based encoders, which recursively concatenate each word with its previous memory, until the whole information of a sentence has been derived. The most common RNN encoders are Long Short-Term Memory Networks (LSTM; \citealp{hochreiter1997long}) and Gated Recurrent Unit~\cite{chung2014empirical}. RNNs have surpassed the performance of traditional baselines in many NLP tasks~\cite{dai2015semi}. There are also convolutional neural network (CNN; \citealp{lecun1989optimal}) based encoders, which concatenate the sentence information by applying multiple convolving filters over the sentence. CNNs have achieved state-of-the-art results on various NLP tasks~\cite{collobert2011natural}. 

To evaluate the quality of the NLI model, the Stanford Natural Language Inference (SNLI; \citealp{bowman2015large}) corpus of 570K sentence pairs was introduced. It serves as a standard benchmark for NLI task. However, most of the sentences in SNLI corpus are short and simple, which limit the room for fine-grained comparisons between models. Currently, a more comprehensive Multi-Genre NLI corpus (MNLI; \citealp{williams2017broad}) of 433K sentence pairs was released, aiming at evaluating large-scale NLI models. Authors gave out some baseline results accompanied by the publish of MNLI corpus, the BiLSTM model achieves an accuracy of 67.5, and the Enhanced Sequential Inference Model~\cite{chen2016enhancing} achieves an accuracy of 72.4.

Among those encoders for NLI task, most of them use word-level embedding, and initialize the weight of the embedding layer with pre-trained word vectors such as GloVe~\cite{pennington2014glove}. The pre-trained word vectors helps the encoders to catch richer semantic information. However, it also has its downside. As the growth of vocabulary size in the modern corpus, there will be more and more out-of-vocabulary (OOV) words that are not presented in the pre-trained word embedding vector. As the word-level embedding is blind to subword information (e.g. morphemes), it leads to high perplexities for those OOV words.

In this paper, we use the BiLSTM model from~\cite{williams2017broad} as the baseline model for the evaluation of the MNLI corpus. To augment the baseline model, firstly, a character-level convolutional neural network (CharCNN; \citealp{kim2016character}) is applied. We use the CharCNN to replace the word embedding layer in the baseline model, which will be computed from the characters of corresponding word. Secondly, the intra attention mechanism introduced by~\cite{yang2016hierarchical} will be applied, to enhance the model with a richer information of substructures of a sentence.

\section{Model Development}
\subsection{BiLSTM Baseline}
The baseline model we used here is introduced by~\cite{williams2017broad} accompanied with the publication of MNLI corpus. It has a 5-layer structure which is shown in Figure 1. 

\begin{figure}[ht]\label{fig:1}
  \centering
  \includegraphics[width=7.5cm]{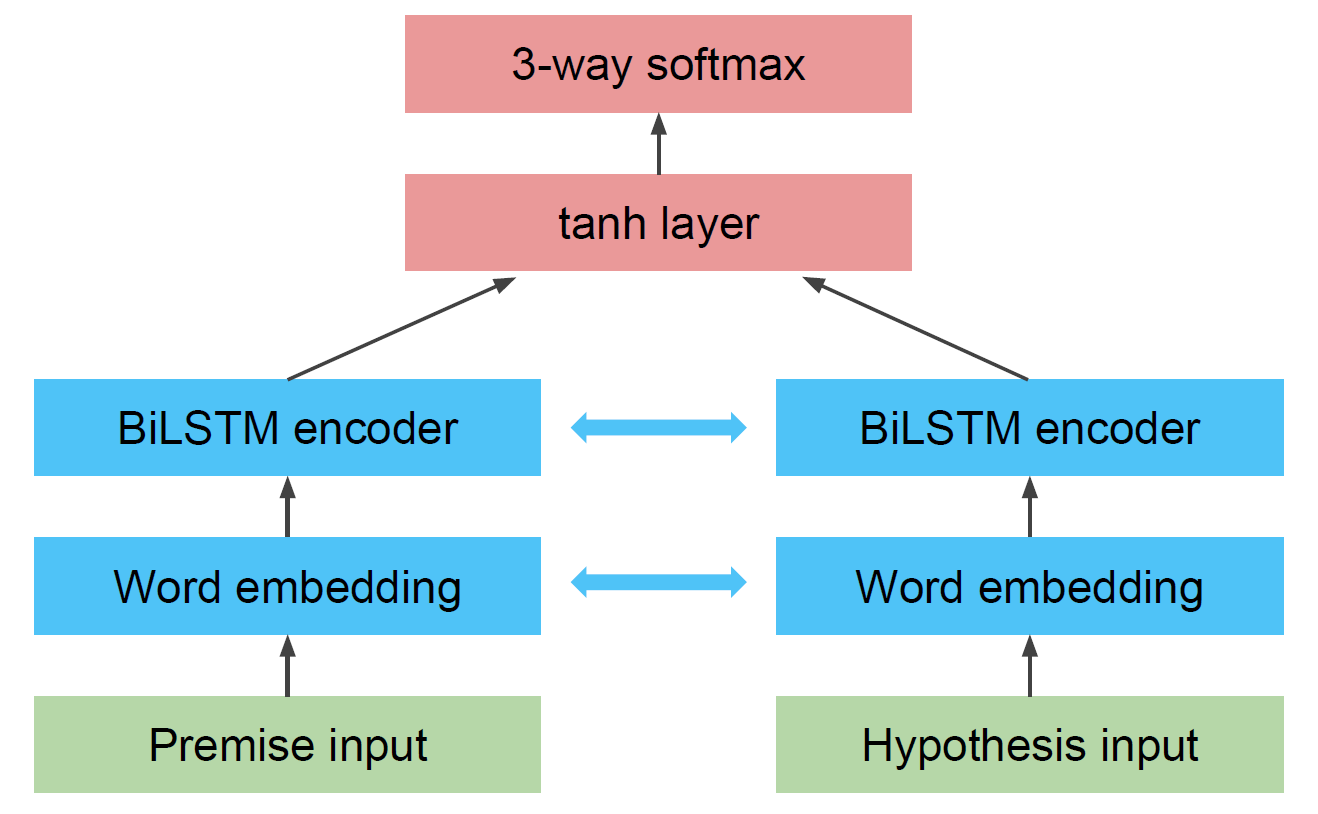}
  \caption{BiLSTM model architecture}
\end{figure}

In the baseline model, a word embedding layer initialized with pre-trained GloVe vectors (840B token version) is implemented to transform the input text into sequence of word vectors. OOV words are initialized randomly. Then, the sentence representation vector $h$ is produced by implementing an average pooling over the BiLSTM hidden states $[h_0,h_1, \cdots ,h_n]$. Finally, the concatenation of encoded premise and hypothesis representation vector is passed through a tanh layer followed by a three-way softmax classifier to attain the label prediction.

\subsection{Character-level Convolutional Neural Network}
In the baseline model, the input $x_t$ to the BiLSTM encoder layer at time $t$ is sequence of pre-trained word embeddings. Those pre-trained word embeddings can boost the performance of the model. However, it is limited to the finite-size of vocabulary. Here we replace the word embedding layer with a character-level convolutional neural network (CharCNN; \citealp{kim2016character}) for language modeling, which also achieved success in machine translation~\cite{Costa-Jussa16}. 

We define the text sentence input as vector ${C^k}\in{R^{d\times{l}}}$, where $k\in K$ is the $k$-th word in a sentence,  $d$ is the dimensionality of character embeddings, $l$ is the length of characters in $k$-th word. Then a set of narrow convolutions between $C^k$ and filter $H$ is applied, followed with a max-over-time (max pooling) as shown in Equation  \ref{eq:1}-\ref{eq:2}. 
\begin{equation} \label{eq:1}
f^k[i]=\tanh(\langle{C^k[\ast, i: i+\omega-1], H}\rangle+b)
\end{equation}
\begin{equation} \label{eq:2}
y^k=\max_{i}f^k[i]
\end{equation}
The concatenation of those max pooling values $y^k$ provides us with a representation vector $y$ of each sentence. Then, a highway network is applied upon $y$, as shown in Equation \ref{eq:3}, where $g$ is a nonlinear transformation, $t=\sigma(W_Ty+b_T)$ is called the transform gate, and $(1-t)$ is called the carry gate. Highway layers allow for training of deep networks by adaptively carrying some dimensions of the input $y$ directly to the output $z$.
\begin{equation} \label{eq:3}
z=t\odot{g(W_Hy+b_H)}+(1-t)\odot{y}
\end{equation}
Experiment conducted by~\cite{kim2016character} has shown that the CNN layer can extract the orthographic features of words  (e.g. \textit{German} and \textit{Germany}). It has also been shown that highway layer is able to encode semantic features that are not discernable from orthography alone. For instance, after highway layers the nearest neighbor word of \textit{you} is \textit{we}, which is orthographically distinct from \textit{you}. 

\subsection{Intra Attention Mechanism}
In the baseline model, the BiLSTM encoder takes an average pooling over all its hidden states to produce a single representation vector of each sentence. However, this has its bottleneck as we intuitively know that not all words (hidden states) contribute equally to the sentence representation. To augment the performance of RNN based encoder, the concept of attention mechanism was introduced by~\cite{bahdanau2014neural} for machine translation. Attention mechanism is a hidden layer which computes a categorical distribution to make a soft-selection over source elements~\cite{kim2017structured}. It has recently demonstrated success on tasks such as parsing text~\cite{vinyals2015grammar}, sentence summarization~\cite{rush2015neural} and also on a wide range of NLP tasks~\cite{cheng2016long}. 

Here we implemented the Intra Attention mechanism introduced by~\cite{yang2016hierarchical} for document classification. We define the hidden states as the output of the BiLSTM encoder as $h_t\in[h_0,h_1, \cdots ,h_n]$, the intra attention is applied upon the hidden states to get the sentence representation vector $h$, specifically, 
\begin{equation} \label{eq:4}
u_t=\tanh({W_\omega h_t+b_\omega})
\end{equation}
\begin{equation} \label{eq:5}
\alpha_t=\frac{exp(u_t^Tu_\omega)}{\sum_{t} exp(u_t^Tu_\omega)}
\end{equation}
\begin{equation} \label{eq:6}
h=\sum_{t} \alpha_t h_t
\end{equation}
It first feed all hidden states $h_t$ through a nonlinearity to get $u_t$ as the hidden representation of $h_t$. Then it uses a $softmax$ function to catch the normalized importance weight matrix $\alpha_t$. After that, the sentence representation vector $h$ is computed by a weighted sum of all hidden states $h_t$ with the weight matrix $\alpha_t$. The context vector $u_\omega$ can be seen as a high-level representation of the importance of informative words. 

\subsection{Character-level Intra Attention Network}
The overall architecture of the Character-level Intra Attention Network (CIAN) is shown in Figure 2. The CIAN model is consisted with 7 layers, of which the first and the last layers are the same with our baseline model. The 4 layers in middle are our augmented layers that has been introduced in this section.

\begin{figure}[ht]\label{fig:2}
  \centering
    \includegraphics[width=7.5cm]{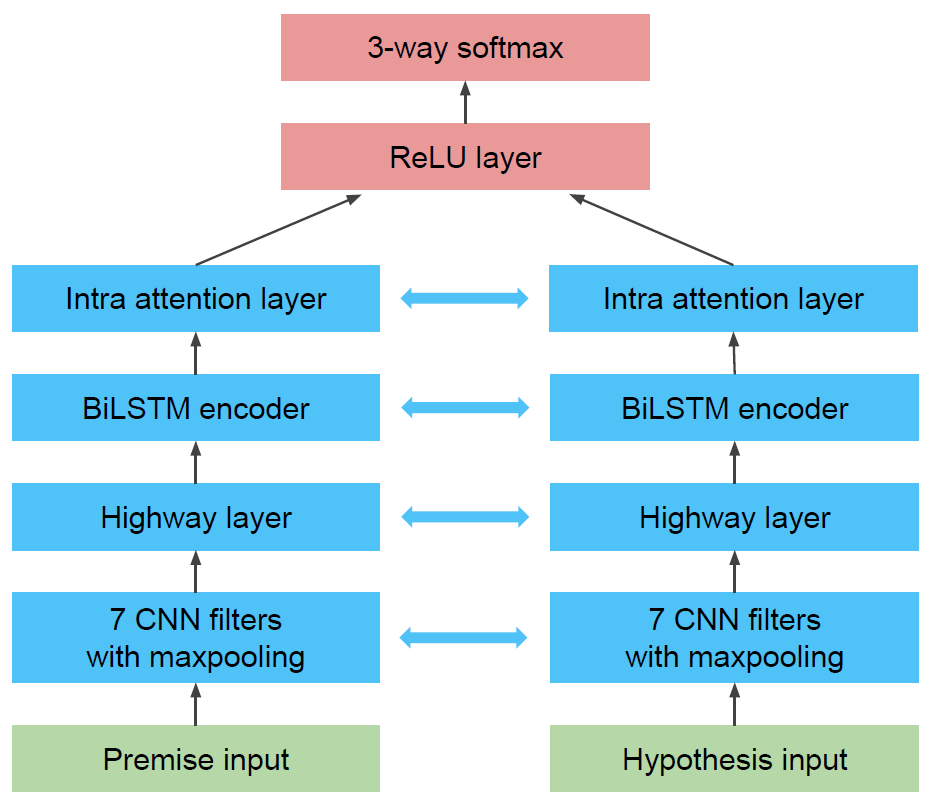}
    \caption{CIAN model architecture}
\end{figure}

The input text is firstly set to lowercase, then it is vectorized according to the tokenization list [abcdefghijklmnopqrstuvwxyz0123456789,;.!?:'"()[]\{\}]. Those characters not in the list are initialized with a vector of zero. After that we use 7 filters in CIAN model's CNN layer. The widths of the CNN filters are $w=[1,2,3,4,5,6,7]$, and the corresponding filters' size are $[\min\{200,50\cdot w\}]$. Two highway layers are implemented following the CNN layer. The attention layer uses weighted sum of all hidden states $h_t$ with the attention weight matrix $\alpha_t$ to encode each sentence into a fixed-length sentence representation vector. Finally a ReLU layer and a three-way softmax classifier use those representation vectors to conduct the prediction.

\section{Experiments}

\subsection{Data}
We evaluated our approach on the Multi-Genre NLI (MNLI) corpus, as a shared task for RepEval 2017 workshop~\cite{nangia2017repeval}. We train our CIAN model on a mixture of MNLI and SNLI corpus, by using a full MNLI training set and a randomly selected 20 percent of the SNLI training set at each epoch. 

\subsection{Hyper Parameters}
The BiLSTM encoder layer use 300D hidden states, thus 600D as it’s a bidirectional encoder. Dropout~\cite{srivastava2014dropout} is implemented with a dropout rate of 0.2 to prevent the model from overfitting. Parameter weights for premise encoder and hypothesis encoder are shared using siamese architecture. The Adam optimizer~\cite{kingma2014adam} is used for training with backpropagation.

The model has been implemented using Keras and we have released the code \footnote{https://github.com/yanghanxy/CIAN}. The training took approximately one hour for one epoch on GeForce GTX TITAN, and we stopped training after 40 epochs as an early stopping regularization. 

\subsection{Result}
We compared the results of CIAN model with the results of BiLSTM model given by~\cite{williams2017broad}. Table 1 shows that the accuracy is improved by 0.9 percent in matched test set, and 0.6 percent in mismatched test set. 

\begin{table}[ht]
\begin{center}
  \begin{tabular}{ l | r  r }
    \hline
    \textbf{Model} & \textbf{Matched} & \textbf{Mismatched} \\ \hline
    BiLSTM & 67.0 & 67.6 \\
    CIAN & 67.9 & 68.2 \\
    \hline
  \end{tabular}
  \caption{Test set accuracies (\%) on MNLI corpus.}
  \label{table:1}
\end{center}
\end{table}

\begin{table}[ht]
\begin{center}
  \resizebox{7.5cm}{!} {
  \begin{tabular}{ l  c | c  c | c  c }
    & \multicolumn{3}{c}{\textbf{Matched}} & \multicolumn{2}{c}{\textbf{Mismatched}} \\ \hline
    \textbf{Tag} && \textbf{BiLSTM} & \textbf{CIAN} & \textbf{BiLSTM} & \textbf{CIAN} \\ \hline
    CONDITIONAL              && 100 & 48  & 100 & 62 \\
    WORD\_OVERLAP            && 50  & 79  & 57  & 62 \\
    NEGATION                 && 71  & 71  & 69  & 70 \\
    ANTO                     && 67  & 82  & 58  & 70 \\
    LONG\_SENTENCE           && 50  & 68  & 55  & 63 \\
    TENCE\_DIFFERNCE         && 64  & 65  & 71  & 72 \\
    ACTIVE/PASSIVE           && 75  & 87  & 82  & 90 \\
    PARAPHRASE               && 78  & 88  & 81  & 89 \\
    QUANTITY/TIME\           && 50  & 47  & 46  & 44 \\
    COREF                    && 84  & 67  & 80  & 72 \\
    QUANTIFIER               && 64  & 63  & 70  & 69 \\
    MODAL                    && 66  & 66  & 64  & 70 \\
    BELIEF                   && 74  & 71  & 73  & 70 \\
    \hline
  \end{tabular}
  }
  \caption{Accuracies (\%) on matched and mismatched expert-tagged development data.}
  \label{table:2}
\end{center}
\end{table}

We conducted error analysis based on expert-tagged development data released by the organizers of RepEval 2017 shared task. The results are shown in Table 2. From the results, it can be seen that the accuracy for WORD\_OVERLAP, LONG\_SENTENCE, ACTIVE/PASSIVE and PARAPHRASE have been improved significantly in both matched and mismatched development set. While the accuracy for CONDITIONAL and COREF haven been decreased in both development set. 

We also conducted visualization on the attention weights $\alpha_t$ of the intra attention layer. By doing so, we we can understand how the model judges the NLI relation between two sentences.

Figure 3 is visualizations of attention weights for 2 sentence pairs, with premise at left and hypothesis at right. Each word is attained with a color block. The darker the color, the greater the attention weight, which means the higher importance contributed to the sentence representation.

\begin{figure}[ht]
\centering
  \includegraphics[width=5.5cm]{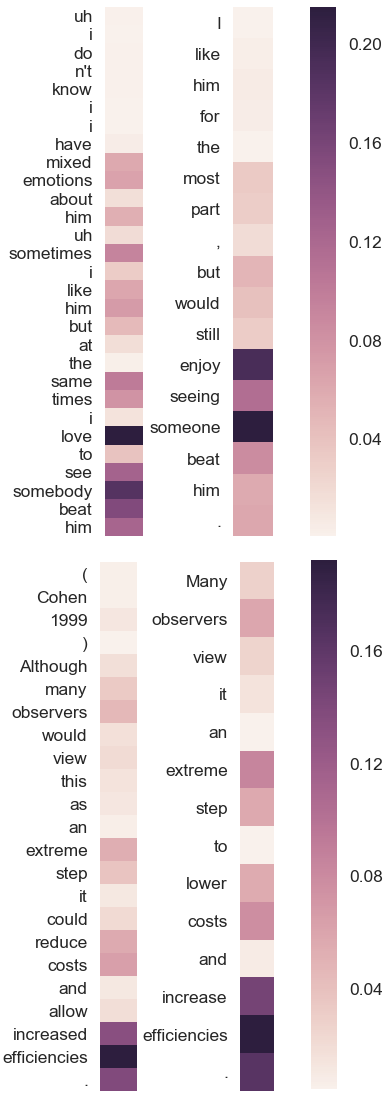}
  \caption{Visualization of attention weights of sentence pair 254941e (top) and 192997e (bottom)}
  \label{fig:3}
\end{figure}

From the Visualization, it could be seen that the model has more attention on words with similar semantic meaning (e.g. \textit{love} and \textit{enjoy}), and the model applies more attention on overlapped words (e.g. \textit{efficiencies} and \textit{efficiencies}). 

\section{Conclusion}
In this paper, we presented a Character-level Intra Attention Network (CIAN) for the task of natural language inference. Experimental results demonstrate that our model slightly outperforms the baseline model upon the MultiNLI corpus. The CharCNN layers helps the model to capture rich semantic and orthographic features. The intra attention layer augment the model's ability to efficiently encode long sentences, and it enhances the models' interpretability by visualizing the attention weights. 

In general, the model presented in this paper is a sequence encoder that do not need any specific pre-processing or outside data like pre-trained word embeddings. Thus, it can be easily applied to other autoencoder architecture tasks such as language modeling, sentiment analysis and question answering.

\section*{Acknowledgments}
This work is supported by the Spanish Ministerio de Econom\'{i}a y Competitividad and Fondo Europeo de Desarrollo Regional through contract TEC2015-69266-P (MINECO/FEDER, UE), by the postdoctoral senior grant Ram\'{o}n y Cajal, and by the China Scholarship Council (CSC) under grant No.201506890038. 

\bibliography{emnlp2017}
\bibliographystyle{emnlp_natbib}

\end{document}